\documentclass[10pt,twocolumn,letterpaper]{article}

\usepackage[pagenumbers]{cvpr} 

\usepackage{graphicx}
\usepackage{amsmath}
\usepackage{amssymb}
\usepackage{booktabs}
\usepackage{colortbl}
\usepackage{multirow}
\usepackage{makecell}
\usepackage{animate}

\usepackage{bm}
\usepackage[accsupp]{axessibility}  

\definecolor{mygray}{RGB}{220, 220, 220}

\newlength{\itemheight} %

\usepackage[pagebackref,breaklinks,colorlinks]{hyperref}

\usepackage[capitalize]{cleveref}
\crefname{section}{Sec.}{Secs.}
\Crefname{section}{Section}{Sections}
\Crefname{table}{Table}{Tables}
\crefname{table}{Tab.}{Tabs.}

\def\cvprPaperID{1492} 
\def\confName{CVPR}
\def\confYear{2023}

\begin{document}

\title{3D Cinemagraphy from a Single Image}

\author{Xingyi Li$^{1,3}$\hspace{0.1in} 
        Zhiguo Cao$^{1}$\hspace{0.1in} 
        Huiqiang Sun$^{1}$\hspace{0.1in}
        Jianming Zhang$^{2}$\hspace{0.1in} 
        Ke Xian$^{3}$\footnotemark[1]~\hspace{0.1in}
        Guosheng Lin$^{3}$\\
$^1$Key Laboratory of Image Processing and Intelligent Control, Ministry of Education \\ School of Artificial Intelligence and Automation, Huazhong University of Science and Technology\\
$^2$Adobe Research\hspace{0.2in} 
$^3$S-Lab, Nanyang Technological University\\
{\tt\small \{xingyi\_li,zgcao,shq1031\}@hust.edu.cn, jianmzha@adobe.com, \{ke.xian,gslin\}@ntu.edu.sg}\\
{\small{\url{https://xingyi-li.github.io/3d-cinemagraphy}}}
\vspace{-2mm}
}


\newcommand{\doanimated}

\twocolumn[{%
\renewcommand\twocolumn[1][]{#1}%
\maketitle
\centering
\setlength{\itemheight}{2.97cm}
\ifdefined\doanimated
\animategraphics[autoplay,loop,trim = 100 0 0 0,height=\itemheight]{20}{figures/teaser/holynski-21/}{0}{59} 
\animategraphics[autoplay,loop,trim = 50 0 0 0,height=\itemheight]{20}{figures/teaser/28-mountain/}{0}{59} 
\animategraphics[autoplay,loop,trim = 50 0 0 0,height=\itemheight]{20}{figures/teaser/11-crayon/}{0}{59} 
\animategraphics[autoplay,loop,height=\itemheight]{20}{figures/teaser/24-diffusion/}{0}{89} 
\else
    \includegraphics[clip,trim = 100 0 0 0,height=\itemheight]{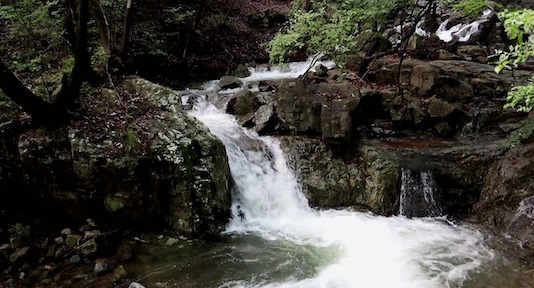} 
    \includegraphics[clip,trim = 50 0 0 0,height=\itemheight]{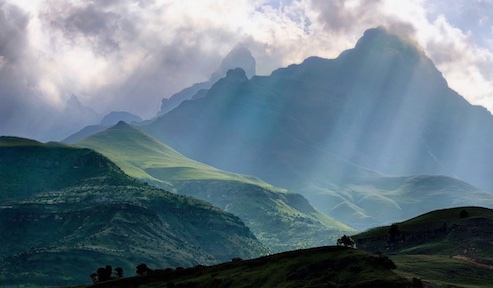}
    \includegraphics[clip,trim = 50 0 0 0,height=\itemheight]{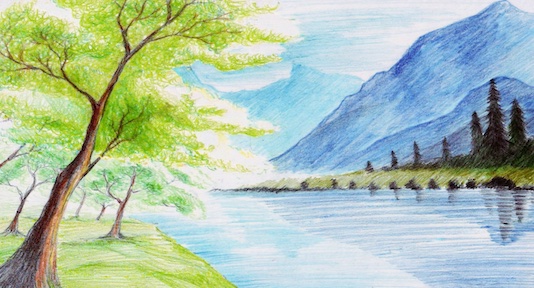}  
    \includegraphics[clip,height=\itemheight]{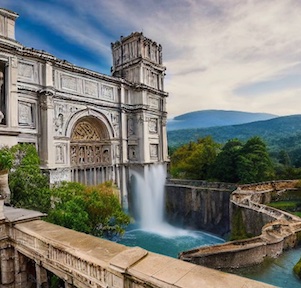} 
\fi
\small{\captionof{figure}{\label{fig:teaser} Given a single still image, our method can synthesize videos with plausible animation of the scene while allowing camera movements. Here, we showcase four 3D cinemagraphs with various camera trajectories. Besides real-world photos (the left two examples), our method can also generalize to paintings (the third one) and synthetic images generated by Stable Diffusion~\cite{rombach2022high} (the rightmost one). To see the effect of 3D cinemagraphy, readers are encouraged to view with Adobe Acrobat or KDE Okular.}}
\vspace{2em}
}]

\renewcommand{\thefootnote}{\fnsymbol{footnote}}
\footnotetext[1]{Corresponding author.}

\begin{abstract}
   We present 3D Cinemagraphy, a new technique that marries 2D image animation with 3D photography. Given a single still image as input, our goal is to generate a video that contains both visual content animation and camera motion. We empirically find that naively combining existing 2D image animation and 3D photography methods leads to obvious artifacts or inconsistent animation. Our key insight is that representing and animating the scene in 3D space offers a natural solution to this task. To this end, we first convert the input image into feature-based layered depth images using predicted depth values, followed by unprojecting them to a feature point cloud. To animate the scene, we perform motion estimation and lift the 2D motion into the 3D scene flow. 
   Finally, to resolve the problem of hole emergence as points move forward, we propose to bidirectionally displace the point cloud as per the scene flow and synthesize novel views by separately projecting them into target image planes and blending the results. 
   Extensive experiments demonstrate the effectiveness of our method. A user study is also conducted to validate the compelling rendering results of our method. 
\end{abstract}

\section{Introduction}
\label{sec:intro}

Nowadays, since people can easily take images using smartphone cameras, the number of online photos has increased drastically. However, with the rise of online video-sharing platforms such as YouTube and TikTok, people are no longer content with static images as they have grown accustomed to watching videos. 
It would be great if we could animate those still images and synthesize videos for a better experience. 
These living images, termed cinemagraphs, have already been created and gained rapid popularity online~\cite{bai2013automatic,yan2017turning}. Although cinemagraphs may engage people with the content for longer than a regular photo, they usually fail to deliver an immersive sense of 3D to audiences. This is because cinemagraphs are usually based on a static camera and fail to produce parallax effects.  
We are therefore motivated to explore ways of animating the photos and moving around the cameras at the same time. 
As shown in Fig.~\ref{fig:teaser}, this will bring many still images to life and provide a drastically vivid experience. 

In this paper, we are interested in making the first step towards \textit{3D cinemagraphy} that allows both realistic animation of the scene and camera motions with compelling parallax effects from a single image. 
There are plenty of attempts to tackle either of the two problems. Single-image animation methods~\cite{endo2019animatinglandscape,Holynski_2021_CVPR,mahapatra2022controllable} manage to produce a realistic animated video from a single image, but they usually operate in 2D space, 
and therefore they cannot create camera movement effects. 
Classic novel view synthesis methods~\cite{debevec1996modeling,buehler2001unstructured,chai2000plenoptic,gortler1996lumigraph,levoy1996light} and recent implicit neural representations~\cite{mildenhall2020nerf,sitzmann2019scene,park2019deepsdf} entail densely captured views as input to render unseen camera perspectives. Single-shot novel view synthesis approaches~\cite{Shih3DP20,jampani2021slide,niklaus20193d,wiles2020synsin} exhibit the potential for generating novel camera trajectories of the scene from a single image.
Nonetheless, these methods usually hypothesize that the observed scene is static without moving elements. 
Directly combining existing state-of-the-art solutions of single-image animation and novel view synthesis yields visual artifacts or inconsistent animation. 

To address the above challenges, we present a novel framework that solves the joint task of image animation and novel view synthesis. This framework can be trained to create 3D cinemagraphs from a single still image. Our key intuition is that handling this new task in 3D space would naturally enable both animation and moving cameras simultaneously. With this in mind, we first represent the scene as feature-based layered depth images (LDIs)~\cite{shade1998layered} and unproject the feature LDIs into a feature point cloud. To animate the scene, we perform motion estimation and lift the 2D motion to 3D scene flow using depth values predicted by DPT~\cite{Ranftl_2021_ICCV}.  
Next, we animate the point cloud 
according to 
the scene flow. To resolve the problem of hole emergence as points move forward, we are inspired by prior works~\cite{Holynski_2021_CVPR,bao2019depth,niklaus2020softmax} and propose a 3D symmetric animation technique to bidirectionally displace point clouds, which can effectively fill in those unknown regions. Finally, we synthesize novel views at time $t$ by rendering point clouds into target image planes and blending the results. 
In this manner, our proposed method can automatically create 3D cinemagraphs from a single 
image. Moreover, our framework is highly extensible, e.g., we can augment our motion estimator with user-defined masks and flow hints for accurate flow estimation and controllable animation.

In summary, our main contributions are:
\begin{itemize}
    \item We propose a new task of creating 3D cinemagraphs from single images. To this end, we propose a novel framework that jointly learns to solve the task of image animation and novel view synthesis in 3D space.
    \item We design a 3D symmetric animation technique to address the hole problem as points move forward. 
    \item Our framework is flexible and customized. We can achieve controllable animation by augmenting our motion estimator with user-defined masks and flow hints.
\end{itemize}

\section{Related Work}
\label{sec:related-work}

\begin{figure*}[htbp]
    \centering
    \includegraphics[width=\textwidth]{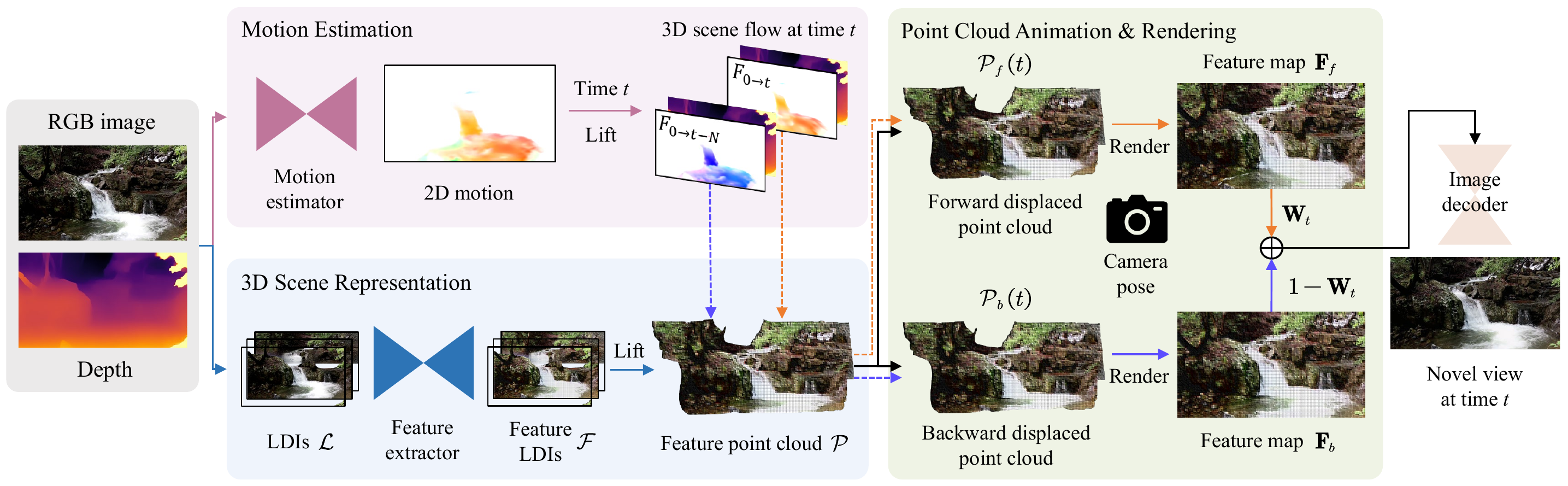}
    \caption{\textbf{An overview of our method.} Given a single still image as input, we first predict a dense depth map. To represent the scene in 3D space, we separate the input image into several layers according to depth discontinuities and apply context-aware inpainting, yielding layered depth images (LDIs) $\mathcal{L}$. We then use a 2D feature extractor to encode 2D feature maps for each inpainted LDI color layer, resulting in feature LDIs $\mathcal{F}$. Subsequently, we lift feature LDIs into 3D space using corresponding depth values to obtain a feature point cloud $\mathcal{P}$. To animate the scene, we estimate a 2D motion field from the input image and apply Euler integration to generate forward and backward displacement fields $F_{0 \rightarrow t}$ and $F_{0 \rightarrow t-N}$. We then augment displacement fields with estimated depth values to obtain 3D scene flow fields. Next, we bidirectionally displace the feature point cloud $\mathcal{P}$ as per the scene flow and separately project them into target image planes to obtain $\bm{\mathrm{F}}_{f}$ and $\bm{\mathrm{F}}_{b}$. Finally, we blend them together and pass the result through our image decoder to synthesize a novel view at time $t$.}
    \label{fig:pipeline}
\end{figure*}

\noindent\textbf{Single-image animation.}
Different kinds of methods have been explored to animate still images. Some works~\cite{chuang2005animating,jhou2015animating} focus on animating certain objects via physical simulation but may not be easily applied to more general cases of in-the-wild photos. Given driving videos as guidance, there are plenty of methods that attempt to perform motion transfer on static objects with either a priori knowledge of moving objects~\cite{chan2019everybody,ren2020deep,siarohin2018deformable,doukas2021headgan,liu2019liquid} or in an unsupervised manner~\cite{siarohin2019animating,siarohin2019first,siarohin2021motion}. 
They entail reference videos to drive the motion of static objects, and thus do not suit our task. Recent advances in generative models have attracted much attention and motivated the community to develop realistic image and video synthesis methods. Many works~\cite{shaham2019singan,logacheva2020deeplandscape,xiong2018learning,liu2021infinite,lin2021infinity} are based on generative adversarial networks (GANs) and operate transformations in latent space to generate plausible appearance changes and movements. Nonetheless, it is non-trial to allow for explicit control over those latent codes and to animate input imagery in a disentangled manner. As diffusion models~\cite{ho2020denoising,sohl2015deep} improve by leaps and bounds, several diffusion-based works~\cite{singer2022make,ho2022video,ho2022imagen} attempt to generate realistic videos from text or images. However, these methods are time-consuming and expensive in terms of computation. 
Here we focus on methods that utilize learned motion priors to convert a still image into an animated video texture~\cite{li2018flow,Holynski_2021_CVPR,endo2019animatinglandscape,mahapatra2022controllable,fan2022simulating}. In particular, Holynski et al.~\cite{Holynski_2021_CVPR} first synthesize the optical flow of the input image via a motion estimation network, then obtain future frames using the estimated flow field. This method renders plausible animation of fluid elements in the input image but suffers from producing camera motions with parallax. 

\noindent\textbf{Novel view synthesis from a single image.}
Novel view synthesis allows for rendering unseen camera perspectives from 2D images and their corresponding camera poses. Recent impressive synthesis results may credit to implicit neural representations~\cite{mildenhall2020nerf,park2019deepsdf,sitzmann2019scene}. Nevertheless, these methods usually assume dense views as input, which is not always available in most cases. Moreover, they focus on the task of interpolation given multiple views rather than extrapolation. As such, we instead turn to methods aiming at handling single input. Among them, a number of works~\cite{yu2021pixelnerf,tucker2020single,tulsiani2018layer,li2021mine,xu2022sinnerf,han2022single,li2022symmnerf} infer the 3D structure of scenes by learning to predict a scene representation from a single image. These methods are usually trained end-to-end but suffer from generalizing to in-the-wild photos. Most relevant to our work are those approaches~\cite{Shih3DP20,niklaus20193d,wiles2020synsin} that apply depth estimation~\cite{Ranftl_2021_ICCV,xian2018monocular,xian2020structure,wang2022less} followed by inpainting occluded regions. For example, 3D Photo~\cite{Shih3DP20} estimates monocular depth maps and uses the representation of layered depth images (LDIs)~\cite{shade1998layered,peng2022mpib}, in which context-aware color and depth inpainting are performed. To enable fine-grained detail modeling, SLIDE~\cite{jampani2021slide} decomposes the scene into foreground and background via a soft-layering scheme. However, unlike our approach, these methods usually assume the scene is static by default, which largely lessens the sense of reality, especially when some elements such as a creek or smoke are also captured in the input image.

\noindent\textbf{Space-time view synthesis.}
Space-time view synthesis is the task of rendering novel camera perspectives for dynamic scenes in terms of space and time~\cite{li2021neural}. Most of the prior works~\cite{bansal20204d,li2022neural,bemana2020x} rely on synchronized multi-view videos as input, which prevents their wide applicability. To mitigate this requirement, many neural rendering approaches~\cite{li2021neural,park2021nerfies,pumarola2021d} manage to show promising space-time view synthesis results from monocular videos. They usually train each new scene independently, and thus cannot directly handle in-the-wild inputs. Most related to our work, 3D Moments~\cite{wang2022_3dmoments} introduces a novel 3D photography effect where cinematic camera motion and frame interpolation are simultaneously performed. However, this method demands near-duplicate photos as input and is unable to control the animation results. Instead, we show that our method can animate still images while enabling camera motion with 3D parallax. Moreover, we can also extend our system so that users are allowed to interactively control how the photos are animated by providing user-defined masks and flow hints.

\section{Method}
\label{sec:method}

\subsection{Overview}
Given a single still image, our goal is to synthesize plausible animation of the scene and simultaneously enable camera motion. The output of our method is a realistic cinemagraph with compelling parallax effects. Fig.~\ref{fig:pipeline} schematically illustrates our pipeline. Our method starts by estimating a motion field and a depth map from the input image. We then separate the RGBD input into several layers as per depth discontinuities and inpaint occluded regions, followed by extracting 2D feature maps for each layer, resulting in feature LDIs~\cite{shade1998layered}. To enable scene animation, we lift the 2D motion to 3D scene flow and unproject feature LDIs into a feature point cloud using their corresponding depth values. Thereafter, we bidirectionally animate the point cloud with scene flow using our 3D symmetric animation technique. We end up rendering them into two animated feature maps and composite the results to synthesize novel views at time $t$.

\subsection{Motion Estimation}
To animate a still image, we wish to estimate the corresponding motion field for the observed scene. Generally, the motion we witness in the real world is extremely complicated as it is time-varying and many events such as occlusion and collision could occur. Intuitively, we could directly adopt prior optical flow estimation methods~\cite{sun2018pwc,teed2020raft,ilg2017flownet,dosovitskiy2015flownet} to accomplish this. However, it is not trivial since they usually take a pair of images as input to compute optical flow. Endo et al.~\cite{endo2019animatinglandscape} instead propose to learn and predict the motion in a recurrent manner, but this kind of approach is prone to large distortions in the long term. To simplify this, we follow Holynski et al.~\cite{Holynski_2021_CVPR} and assume that a time-invariant and constant-velocity motion field, termed Eulerian flow field, can well approximate the bulk of real-world motions, e.g., water, smoke, and clouds. Formally, we denote $M$ as the Eulerian flow field of the scene, which suggests that 
\begin{equation}
    F_{t \rightarrow{} t+1}(\cdot) = M(\cdot),
\end{equation}
where $F_{t \rightarrow{} t+1}(\cdot)$ represents the optical flow map from frame $t$ to frame $t+1$. This defines how each pixel in the current frame will move in the future. Specifically, we can obtain the next frame via Euler integration:
\begin{equation}
    \bm{\mathrm{x}}_{t+1} = \bm{\mathrm{x}}_{t} + M(\bm{\mathrm{x}}_{t}),
\end{equation}
where $\bm{\mathrm{x}}_{t}$ represents the coordinates of a pixel $\bm{\mathrm{x}}_{t}$ at time $t$. Since the optical flow between consecutive frames is identical, we can easily deduce the displacement field by recursively applying: 
\begin{equation}
    F_{0 \rightarrow t}(\bm{\mathrm{x}}_{0}) = F_{0 \rightarrow t-1}(\bm{\mathrm{x}}_{0}) + M(\bm{\mathrm{x}}_{0} + F_{0 \rightarrow t-1}(\bm{\mathrm{x}}_{0})), 
    \label{eq:displacement-field}
\end{equation}
where $F_{0 \rightarrow t}(\cdot)$ denotes the displacement field from time $0$ to time $t$, which describes the course of each pixel in the input image across future frames. 
To estimate the Eulerian flow field, we adopt an image-to-image translation network as our motion estimator, which is able to map an RGB image to the optical flow.

\subsection{3D Scene Representation}
One common disadvantage of previous single-image animation methods~\cite{li2018flow,Holynski_2021_CVPR,endo2019animatinglandscape} is that they usually operate in 2D space via a deep image warping technique, which prevents them from creating parallax effects. Instead, to enable camera motion, we propose to lift our workspace into 3D and thus resort to 3D scene representation. 

We start by estimating the underlying geometry of the scene using the state-of-the-art monocular depth estimator DPT~\cite{Ranftl_2021_ICCV}, which can predict reasonable dense depth maps for in-the-wild photos. Following Wang et al.~\cite{wang2022_3dmoments}, we then convert the RGBD input into an LDI representation~\cite{shade1998layered} by separating it into several layers as per depth discontinuities and inpainting occluded regions. Specifically, we first divide the depth range of the source depth map into multiple intervals using agglomerative clustering~\cite{maimon2005data}, followed by creating layered depth images $\mathcal{L}=\{ \bm{\mathrm{C}}_{l}, \bm{\mathrm{D}}_{l} \}_{l=1}^{L}$. 
Next, we inpaint occluded regions of each color and depth layer by applying the pretrained inpainting model from 3D Photo~\cite{Shih3DP20}. To improve rendering quality and reduce artifacts, we also introduce a 2D feature extraction network to encode 2D feature maps for each inpainted LDI color layer, resulting in feature LDIs $\mathcal{F}=\{ \bm{\mathrm{F}}_{l}, \bm{\mathrm{D}}_{l} \}_{l=1}^{L}$. 
Finally, in order to enable animation in 3D space, we unproject feature LDIs into 3D via their corresponding inpainted depth layers, yielding a feature point cloud $\mathcal{P} = \{ (\bm{\mathrm{X}}_{i}, \bm{\mathrm{f}}_{i}) \}$, where $\bm{\mathrm{X}}_{i}$ and $\bm{\mathrm{f}}_{i}$ are 3D coordinates and the feature vector for each 3D point respectively.

\subsection{Point Cloud Animation and Rendering}
We now have the estimated displacement fields $F_{0 \rightarrow t}$ and the feature point cloud $\mathcal{P}$. Our next step is to animate this point cloud over time. To bridge the gap between 2D displacement fields and 3D scene representation, we first augment the displacement fields with estimated depth values to lift them into 3D scene flow. In other words, we now have a function of time $t$ and the coordinates of a 3D point that returns a corresponding 3D translation vector that can shift this 3D point accordingly. Thus, for time $t$, we then move each 3D point by computing its destination as its original position plus a corresponding 3D translation vector, i.e., $\mathcal{P}(t) = \{ (\bm{\mathrm{X}}_{i}(t), \bm{\mathrm{f}}_{i}) \}$. Intuitively, this process indeed animates the point cloud from one time to another. However, we empirically find that as points move forward, increasingly large holes emerge. This frequently happens when points leave their original locations without any points filling in those unknown regions. 

\begin{figure}[t]
    \centering
    \includegraphics[width=0.46\textwidth]{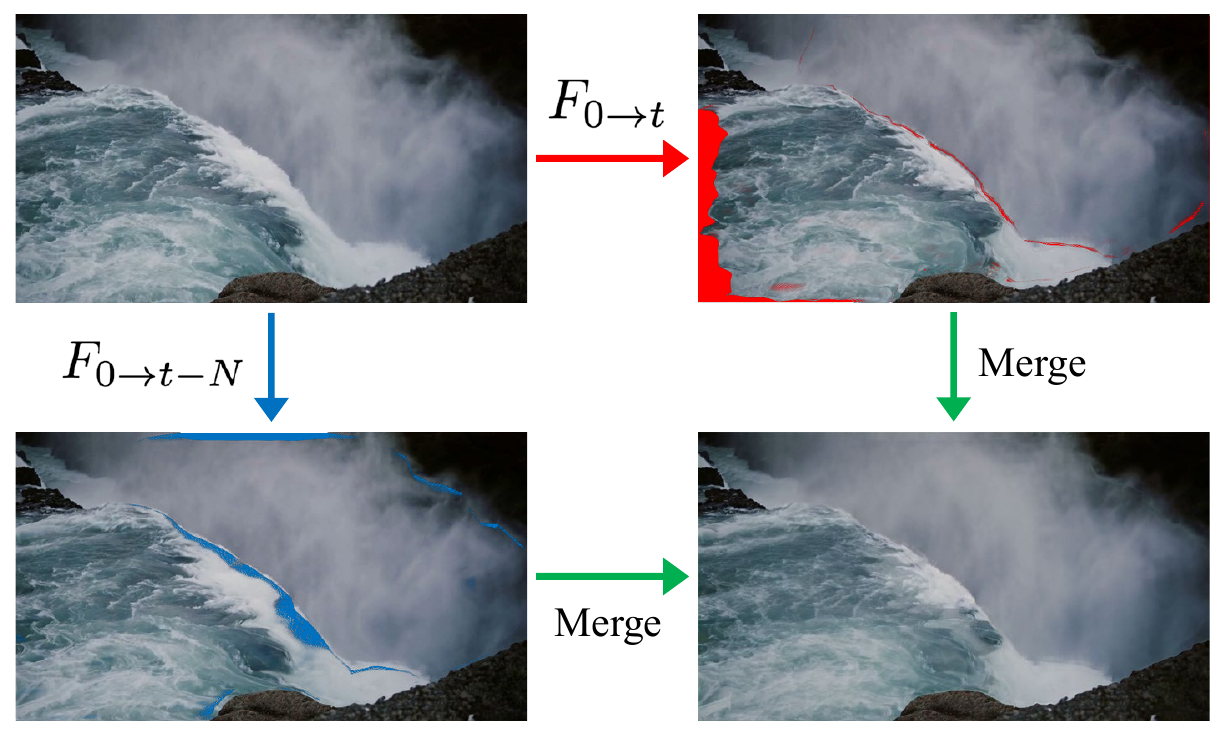}
    \caption{\textbf{3D symmetric animation.} To address the hole issue, we borrow textural information from the point cloud that moves in the opposite direction and integrate both of the animated point clouds to feasibly fill in the missing regions (the {\color{red} red} and {\color{blue} blue} regions). }
    \vspace{-4mm}
    \label{fig:3d-symmetric-animation}
\end{figure}

\noindent\textbf{3D symmetric animation.}
To resolve this, inspired by prior works~\cite{Holynski_2021_CVPR,niklaus2020softmax,bao2019depth}, we propose a 3D symmetric animation technique that leverages bidirectionally displaced point clouds to complement each other. With 3D symmetric animation, we can borrow textural information from point clouds that move in the opposite direction and integrate both of the animated point clouds to feasibly fill in missing regions. Specifically, we directly replace the original Eulerian flow field $M$ with $-M$ and recursively apply Eq.~\eqref{eq:displacement-field} to generate a reversed displacement field. Similarly, we then lift this 2D displacement field to obtain inverse scene flow, which is employed to produce point clouds with backward movements. As illustrated in Fig.~\ref{fig:3d-symmetric-animation}, for time $t$, to fill in holes, we respectively apply $F_{0 \rightarrow t}$ and $F_{0 \rightarrow t-N}$ to draw associated scene flow fields and use them to move the point cloud, resulting in $\mathcal{P}_{f}(t) = \{ (\bm{\mathrm{X}}_{i}^{f}(t), \bm{\mathrm{f}}_{i}) \}$ and $\mathcal{P}_{b}(t) = \{ (\bm{\mathrm{X}}_{i}^{b}(t), \bm{\mathrm{f}}_{i}) \}$, where $N$ is the number of frames.

\noindent\textbf{Neural rendering.}
We now have two bidirectionally animated feature point clouds. Our final step is to render them into animated feature maps and composite the results for synthesizing novel views at time $t$. In particular, given camera poses and intrinsics, we use a differentiable point-based renderer~\cite{wiles2020synsin} to splat feature point clouds $\mathcal{P}_{f}(t) = \{ (\bm{\mathrm{X}}_{i}^{f}(t), \bm{\mathrm{f}}_{i}) \}$ and $\mathcal{P}_{b}(t) = \{ (\bm{\mathrm{X}}_{i}^{b}(t), \bm{\mathrm{f}}_{i}) \}$ separately into the target image plane. This process yields 2D feature maps $\bm{\mathrm{F}}_{f}$ and $\bm{\mathrm{F}}_{b}$ along with depth maps $\bm{\mathrm{D}}_{f}$, $\bm{\mathrm{D}}_{b}$ and alpha maps $\bm{\mathrm{\alpha}}_{f}$, $\bm{\mathrm{\alpha}}_{b}$. Next, we wish to fuse $\bm{\mathrm{F}}_{f}$ and $\bm{\mathrm{F}}_{b}$ into one feature map $\bm{\mathrm{F}}_{t}$. Inspired by prior work~\cite{wang2022_3dmoments}, our intuition is three-fold: 1) to enable endless and seamless looping, we should assign the weight of the two feature maps based on time so as to guarantee that the first and last frame of the synthesized video are identical; 2) the weight map should favor pixel locations with smaller depth values, in the sense that it is impossible to see objects behind those objects closer to the eye; 3) to avoid missing regions as much as possible, we should greatly increase the contribution of those pixel locations that can fill in holes. With this in mind, we formulate the weight map as follows:
\begin{equation}
    \bm{\mathrm{W}}_{t} = \frac{(1 - \frac{t}{N}) \cdot \bm{\mathrm{\alpha}}_{f} \cdot e^{-\bm{\mathrm{D}}_{f}} }{(1 - \frac{t}{N}) \cdot \bm{\mathrm{\alpha}}_{f} \cdot e^{-\bm{\mathrm{D}}_{f}} + \frac{t}{N} \cdot \bm{\mathrm{\alpha}}_{b} \cdot e^{-\bm{\mathrm{D}}_{b}}},
\end{equation}
where $N$ is the number of frames. Therefore, we can integrate $\bm{\mathrm{F}}_{f}$ and $\bm{\mathrm{F}}_{b}$ via:
\begin{equation}
    \bm{\mathrm{F}}_{t} = \bm{\mathrm{W}}_{t} \cdot \bm{\mathrm{F}}_{f} + (1 - \bm{\mathrm{W}}_{t}) \cdot \bm{\mathrm{F}}_{b}.
\end{equation}
We also obtain the merged depth map $\bm{\mathrm{D}}_{t}$: 
\begin{equation}
    \bm{\mathrm{D}}_{t} = \bm{\mathrm{W}}_{t} \cdot \bm{\mathrm{D}}_{f} + (1 - \bm{\mathrm{W}}_{t}) \cdot \bm{\mathrm{D}}_{b}.
\end{equation}
Finally, we employ an image decoder network to map the 2D feature map $\bm{\mathrm{F}}_{t}$ and depth map $\bm{\mathrm{D}}_{t}$ to a novel view at time $t$. Repeating this method, we are able to synthesize a realistic cinemagraph with compelling parallax effects.

\subsection{Training}
This section describes our training scheme. In general, we train our image-to-image translation network, 2D feature extraction network, and image decoder network in a two-stage manner. 

\noindent\textbf{Training dataset.}
We use the training set from Holynski et al.~\cite{Holynski_2021_CVPR} as our training dataset. This dataset comprises short video clips of fluid motion that are extracted from longer stock-footage videos. We use the first frames of each video clip and the corresponding ground truth motion fields estimated by a pretrained optical flow network~\cite{sun2018pwc} as motion estimation pairs to train our motion estimation network. To develop animation ability, we randomly sample training data from fluid motion video clips. For novel view synthesis training, we require multi-view supervision of the same scene, which is not available in the training set. Instead, we use 3D Photo~\cite{Shih3DP20} to generate pseudo ground truth novel views for training. 

\noindent\textbf{Two-stage training.}
Our model is trained in a two-stage manner. Specifically, we first train our motion estimation network using motion estimation pairs. To train the motion estimation network, we minimize GAN loss, GAN feature matching loss~\cite{salimans2016improved}, and endpoint error as follows:
\begin{equation}
    \mathcal{L}_{Motion} = \mathcal{L}_{GAN} + 10 \mathcal{L}_{FM} + \mathcal{L}_{EPE}. 
\end{equation}
In the second stage, we freeze the motion estimation network and train the feature extraction network and image decoder network. Our model simultaneously learns to render novel views and animate scenes. For novel view synthesis, we set $t=0$ and use pseudo ground truth novel views to supervise our model. We randomly sample target viewpoints of scenes and require the model to synthesize them. For animation, we train our model on training triplets (start frame, middle frame, end frame) sampled from fluid motion video clips. In particular, we render the middle frame from both directions using $F_{0 \rightarrow t}$ and $F_{0 \rightarrow t-N}$ without changing the camera poses and intrinsics. Besides GAN loss and GAN feature matching loss~\cite{salimans2016improved}, we also enforce VGG perceptual loss~\cite{johnson2016perceptual,zhang2018unreasonable} and $l_{1}$ loss between synthesized and ground truth images. The overall loss is as follows:
\begin{equation}
    \mathcal{L}_{Animation} = \mathcal{L}_{GAN} + 10 \mathcal{L}_{FM} + \mathcal{L}_{l_1} + \mathcal{L}_{VGG}.
\end{equation}

\section{Experiments}
\label{sec:exp}

\subsection{Implementation Details}
Our motion estimator is a U-Net~\cite{ronneberger2015u} based generator with $16$ convolutional layers, and we replace Batch Normalization with SPADE~\cite{park2019semantic}. For the feature extraction network and image decoder network, we follow the network architectures from Wang et al.~\cite{wang2022_3dmoments}. We adopt the multi-scale discriminator used in SPADE~\cite{park2019semantic} during training.

Our model is trained using the Adam optimizer~\cite{kingma2014adam}. We conduct all experiments on a single NVIDIA GeForce RTX 3090 GPU. We train the motion estimation network for around $120k$ iterations with a batch size of $16$. We set the generator learning rate to $5 \times 10^{-4}$ and the discriminator learning rate to $2 \times 10^{-3}$. For the animation training stage, we train the feature extraction network and image decoder network for around $250k$ iterations with a learning rate starting at $1 \times 10^{-4}$ and then decaying exponentially. 

\subsection{Baselines}
\label{sec:baselines}
In principle, to evaluate our method, we are required to compare it against current state-of-the-art models. However, to our knowledge, we are the first to tackle the novel task of synthesizing a realistic cinemagraph with compelling parallax effects from a single image. As a result, we cannot directly compare to previous works. Instead, we consider forming the following baselines to verify the superiority of our method:

\noindent\textbf{2D animation $\rightarrow$ novel view synthesis.}
One might consider 2D image animation $\rightarrow$ single-shot novel view synthesis: first employing a 2D image animation method, then a single-shot novel view synthesis method. Specifically, we first adopt a state-of-the-art image animation method~\cite{Holynski_2021_CVPR} to produce an animated looping video. We then apply DPT~\cite{Ranftl_2021_ICCV} to estimate geometry and utilize 3D Photo~\cite{Shih3DP20} to generate novel views for each frame. 

\noindent\textbf{Novel view synthesis $\rightarrow$ 2D animation.}
It also appears to be feasible that we first render novel views of scenes by 3D Photo~\cite{Shih3DP20} and then use the image animation method~\cite{Holynski_2021_CVPR} to animate each viewpoint. Note that motion estimation should be performed for each frame as viewpoints have changed. However, we empirically find that this usually results in varying motion fields across the video. To mitigate this, 
we further propose using the moving average technique to smooth estimated motions for each frame. This results in novel view synthesis $\rightarrow$ 2D animation + MA.

\noindent\textbf{Naive point cloud animation.}
Intuitively, we may also consider directly unprojecting pixels into 3D space and subsequently moving and rendering the RGB point cloud. Specifically, given a single input image, we first predict the depth map using DPT~\cite{Ranftl_2021_ICCV} and estimate 2D optical flow. We then lift the pixels and optical flow into 3D space to form RGB point clouds and scene flow. Finally, we animate RGB point clouds over time according to the scene flow and project these point clouds into target viewpoints. This baseline also faces a similar issue: as time goes by, large holes gradually appear. One might also employ our 3D symmetric animation technique to further enhance this baseline, i.e., naive point cloud animation + 3DSA.

\subsection{Results}

\begin{figure*}[htbp]
    \centering
    \includegraphics[width=\textwidth]{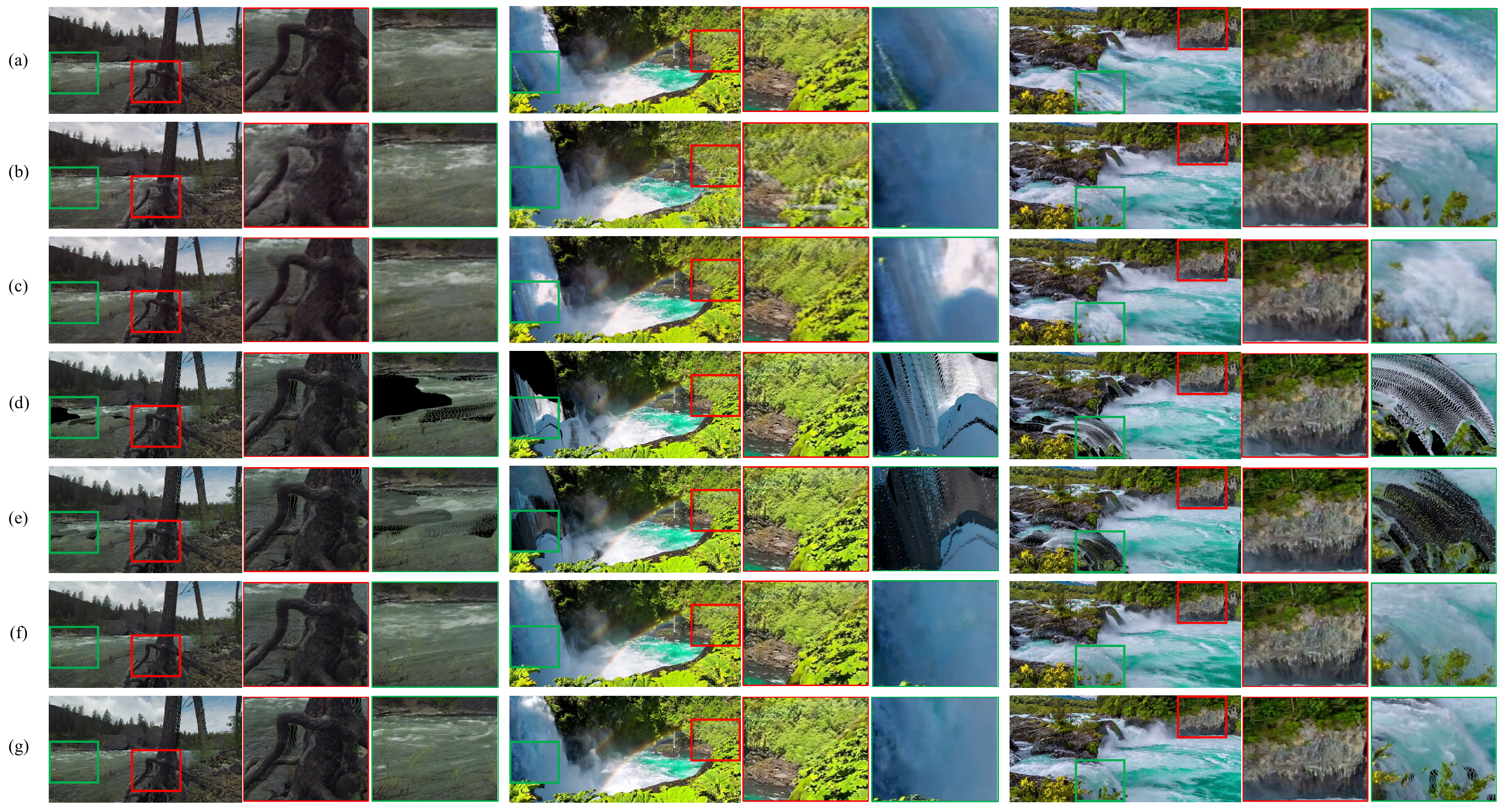}
    \caption{\textbf{Qualitative comparisons against all baselines on the validation set from Holynski et al.~\cite{Holynski_2021_CVPR}.} Our method produces compelling results while other comparative alternatives suffer from visual artifacts. (a) 2D animation~\cite{Holynski_2021_CVPR} $\rightarrow$ novel view synthesis~\cite{Shih3DP20}, (b) novel view synthesis~\cite{Shih3DP20} $\rightarrow$ 2D animation~\cite{Holynski_2021_CVPR}, (c) novel view synthesis~\cite{Shih3DP20} $\rightarrow$ 2D animation~\cite{Holynski_2021_CVPR} + moving average, (d) naive point cloud animation, (e) naive point cloud animation + 3D symmetric animation, (f) our method, and (g) pseudo ground truth. }
    \vspace{-2mm}
    \label{fig:qualitative}
\end{figure*}

\begin{table}
    \footnotesize
    \centering 
    \caption{{\bf Quantitative comparisons against all baselines on the validation set from Holynski et al.~\cite{Holynski_2021_CVPR}.} The better approach favors higher PSNR and SSIM but lower LPIPS. The best performance is in \textbf{bold}. 
    }
    \resizebox{\linewidth}{!}{
        \renewcommand\arraystretch{0.85}
		\begin{tabular}{lccc}
            \toprule
            Method & PSNR$\uparrow$ & SSIM$\uparrow$ & LPIPS$\downarrow$ \\
            \midrule
            2D Anim.~\cite{Holynski_2021_CVPR} $\rightarrow$ NVS~\cite{Shih3DP20} & 21.12 & 0.633 & 0.286 \\
            NVS~\cite{Shih3DP20} $\rightarrow$ 2D Anim.~\cite{Holynski_2021_CVPR} & 21.97 & 0.697 & 0.276 \\
            NVS~\cite{Shih3DP20} $\rightarrow$ 2D Anim.~\cite{Holynski_2021_CVPR} + MA & 22.47 & 0.718 & 0.261 \\
            Naive PC Anim. & 19.46 & 0.647 & 0.243 \\
            Naive PC Anim. + 3DSA & 20.49 & 0.660 &  0.237 \\
            Ours & \textbf{23.33} & \textbf{0.776} & \textbf{0.197} \\
            \bottomrule
		\end{tabular}
    }
    \vspace{-2mm}
    \label{tab:quantitative}
\end{table}

\begin{figure}[t]
    \centering
    \includegraphics[width=0.47\textwidth]{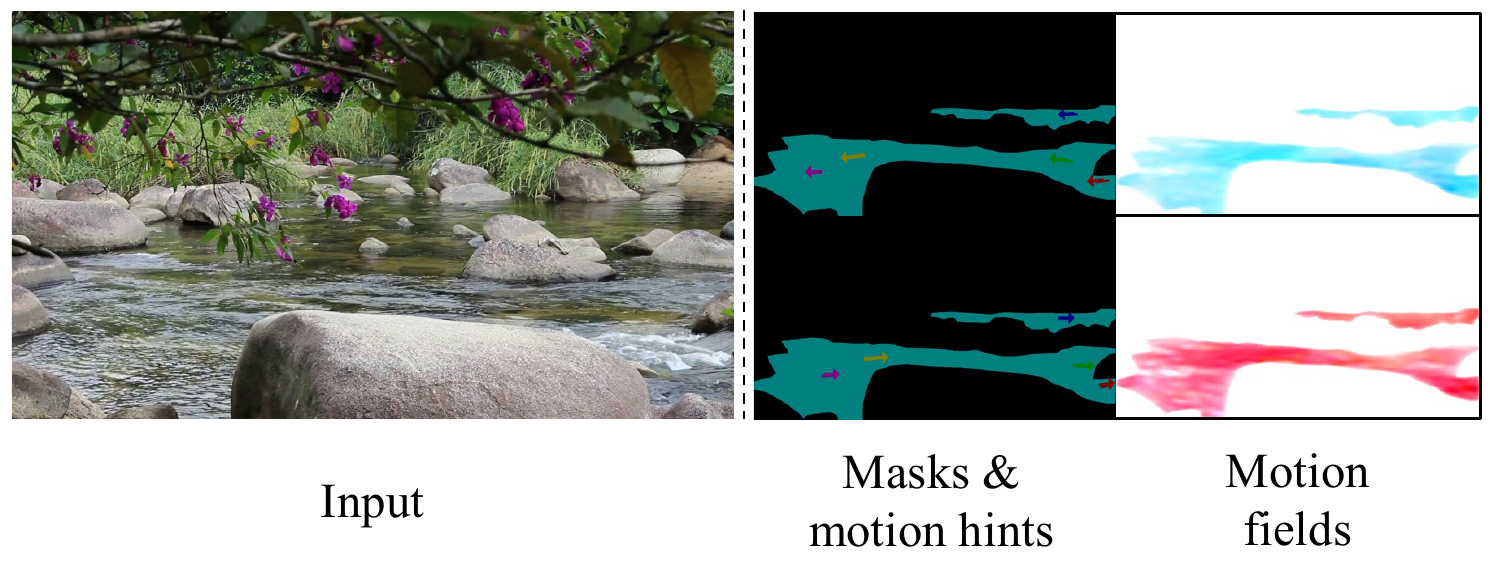}
    \caption{\textbf{Controllable animation.} 
    By changing the masks and motion hints, our method can interactively control the animation.
    }
    \vspace{-5mm}
    \label{fig:controllable}
\end{figure}

\noindent\textbf{Evaluation dataset.}
Since Holynski et al.~\cite{Holynski_2021_CVPR} only provide a single image for each scene in the test set, we use the validation set from Holynski et al.~\cite{Holynski_2021_CVPR} to evaluate our method and baselines. The validation set consists of $31$ unique scenes with $162$ samples of ground truth video clips captured by static cameras. 

\noindent\textbf{Experimental setup.}
For evaluation, we render novel views of the ground truth videos in $4$ different trajectories, resulting in $240$ ground truth frames for each sample. This process does not involve inpainting, thus ground truth frames may contain holes. Only considering valid pixels when calculating metrics, we compare the predicted images with the ground truth frames at the same time and viewpoint. For a fair comparison, all methods utilize the depth maps estimated by DPT~\cite{Ranftl_2021_ICCV}. Since we focus on comparing rendering quality, all methods use ground truth optical flows, except that NVS~\cite{Shih3DP20} $\rightarrow$ 2D Anim.~\cite{Holynski_2021_CVPR} and NVS~\cite{Shih3DP20} $\rightarrow$ 2D Anim.~\cite{Holynski_2021_CVPR} + MA have to estimate optical flows for each frame apart from the first frame. We adopt PSNR, SSIM, and LPIPS~\cite{zhang2018unreasonable} as our evaluation metrics. 

\noindent\textbf{Quantitative comparisons.}
As shown in Table~\ref{tab:quantitative}, our method outperforms all baselines across all metrics by a large margin. This result implies that our method achieves better perceptual quality and produces more realistic renderings, which demonstrates the superiority and effectiveness of our method. 

\noindent\textbf{Qualitative comparisons.}
We showcase the visual comparisons in Fig.~\ref{fig:qualitative}. One can observe that our method presents photorealistic results while other comparative baselines produce more or less visual artifacts. 2D Anim.~\cite{Holynski_2021_CVPR} $\rightarrow$ NVS~\cite{Shih3DP20} intends to generate stripped flickering artifacts. This is because 2D Anim.~\cite{Holynski_2021_CVPR} $\rightarrow$ NVS~\cite{Shih3DP20} predicts the depth map for each animated frame, leading to frequent changes in the 3D structure of the scene and inconsistent inpainting. NVS~\cite{Shih3DP20} $\rightarrow$ 2D Anim.~\cite{Holynski_2021_CVPR} and NVS~\cite{Shih3DP20} $\rightarrow$ 2D Anim.~\cite{Holynski_2021_CVPR} + MA show jelly-like effects as optical flow should be estimated for each novel view. This results in varying motion fields across the video and thus inconsistent animation. Although Naive PC Anim. and Naive PC Anim. + 3DSA also lift the workspace into 3D, they are often prone to produce noticeable holes inevitably. One reason for this is that they do not perform inpainting. Note that some artifacts are difficult to observe when only scanning static figures. 

\begin{table}
    \footnotesize
    \centering 
    \caption{{\bf User study.} Pairwise comparison results indicate that users prefer our method as more realistic and immersive.}
    \resizebox{\linewidth}{!}{
        \renewcommand\arraystretch{0.85}
		\begin{tabular}{lc}
            \toprule
            Comparison & Human preference \\
            \midrule
            2D Anim.~\cite{Holynski_2021_CVPR} $\rightarrow$ NVS~\cite{Shih3DP20} / Ours & 12.5\% / \textbf{87.5\%}  \\
            NVS~\cite{Shih3DP20} $\rightarrow$ 2D Anim.~\cite{Holynski_2021_CVPR} / Ours & 3.9\% / \textbf{96.1\%} \\
            NVS~\cite{Shih3DP20} $\rightarrow$ 2D Anim.~\cite{Holynski_2021_CVPR} + MA / Ours & 6.1\% / \textbf{93.9\%} \\
            Naive PC Anim. / Ours & 7.6\% / \textbf{92.4\%} \\
            Naive PC Anim. + 3DSA / Ours & 8.6\% / \textbf{91.4\%} \\
            3D Photo~\cite{Shih3DP20} / Ours & 10.5\% / \textbf{89.5\%} \\
            Holynski et al.~\cite{Holynski_2021_CVPR} / Ours & 29.9\% / \textbf{70.1\%} \\
            \bottomrule
		\end{tabular}
    }
    \vspace{-2mm}
    \label{tab:user-study}
\end{table}

\begin{table}
    \footnotesize
    \centering 
    \caption{{\bf Ablation study on each component of our method.}}
    \resizebox{1.0\linewidth}{!}{
        \renewcommand\arraystretch{0.85}
		\begin{tabular}{lccc}
            \toprule
             & PSNR$\uparrow$ & SSIM$\uparrow$ & LPIPS$\downarrow$ \\
            \midrule
            w/o features & 21.50 & 0.674 & 0.228 \\
            w/o inpainting & 22.86 & 0.763 & 0.216 \\
            w/o 3D symmetric animation & 22.99 & 0.768 & 0.199 \\
            Full model& \textbf{23.33} & \textbf{0.776} & \textbf{0.197} \\
            \bottomrule
		\end{tabular}
    }
    \vspace{-4mm}
    \label{tab:ablation}
\end{table}

\noindent\textbf{Controllable animation.}
Our method is able to create 3D cinemagraphs from a single image automatically. Further, we show that our framework is also highly extensible. For example, 
we can involve masks and flow hints as extra inputs to augment our motion estimator. This brings two advantages: (1) more accurate flow estimation; (2) interactive and controllable animation. 
As shown in Fig.~\ref{fig:controllable}, we can control the animation of the scene by providing various masks and motion hints to obtain different motion fields. 

\begin{figure*}[htbp]
    \centering
    \includegraphics[width=\textwidth]{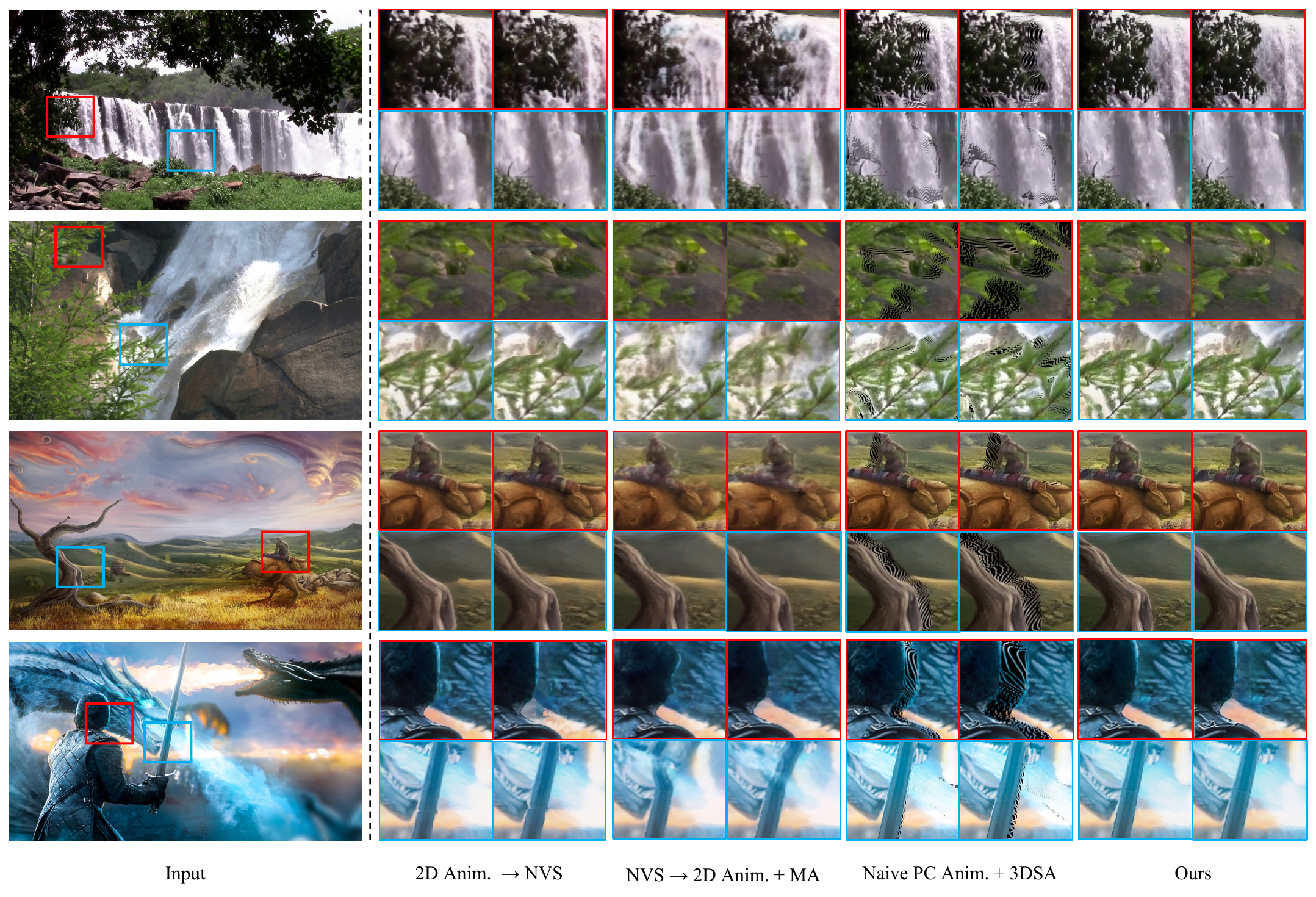}
    \caption{\textbf{Visual comparisons on the test set from Holynski et al.~\cite{Holynski_2021_CVPR} and in-the-wild photos.} Our method consistently produces more realistic rendering with fewer visual artifacts as opposed to other baselines. }
    \vspace{-4mm}
    \label{fig:in-the-wild}
\end{figure*}

\noindent\textbf{Generalizing on in-the-wild photos.}
To further demonstrate the generalization of our method, we also test our method on in-the-wild photos. We first create cinemagraphs with camera motions on the test set from Holynski et al.~\cite{Holynski_2021_CVPR}, where, for each scene, only a single image is provided. We then select some online images at random to test our method. 
To accurately estimate motion fields, we provide masks and flow hints as extra inputs to our motion estimator. 
As shown in Fig.~\ref{fig:in-the-wild}, our method produces reasonable results for in-the-wild inputs while other comparative alternatives yield visual artifacts or inconsistent animation. 

\subsection{User Study}
We further conduct a user study to investigate how our method performs in the view of humans when compared with all baselines, 3D Photo~\cite{Shih3DP20}, and Holynski et al.~\cite{Holynski_2021_CVPR}. Specifically, we collect $50$ photos from the test set of Holynski et al.~\cite{Holynski_2021_CVPR} and the Internet. We use different approaches to generate videos with identical settings. During the study, we show each participant an input image and two animated videos generated by our method and a randomly selected approach in random order. $108$ volunteers are invited to choose the method with better perceptual quality and realism, or none if it is hard to judge. We report the results in Table~\ref{tab:user-study}, which points out that our method surpasses alternative methods by a large margin in terms of the sense of reality and immersion.

\subsection{Ablation Study}
To validate the effect of each component, we conduct an ablation study on the validation set from Holynski et al.~\cite{Holynski_2021_CVPR} and show the results in Table~\ref{tab:ablation}.
One can observe: i) 3D symmetric animation technique matters because it allows us to leverage bidirectionally displaced point clouds to complement each other and feasibly fill in missing regions; ii) introducing inpainting when constructing 3D geometry can improve the performance as this allows our model to produce plausible structures around depth discontinuities and fill in holes; iii) switching from directly using RGB colors to features in 3D scene representation significantly improves the rendering quality and reduces artifacts.

\section{Conclusion}
\label{sec:conclusion}

In this paper, we introduce a novel task of creating 3D cinemagraphs from single images. To this end, we present a simple yet effective method that makes a connection between image animation and novel view synthesis. 
We show that our method produces plausible animation of the scene while allowing camera movements. Our framework is flexible and customized. For accurate motion estimation and controllable animation, we can further include masks and flow hints as extra input for the motion estimator. Therefore, users can control how the scene is animated. Furthermore, our method generalizes well to in-the-wild photos, even like paintings or synthetic images generated by diffusion models. We conduct extensive experiments to verify the effectiveness and superiority of our method. 
A user study also demonstrates that our method generates realistic 3D cinemagraphs. 
We hope that our work can bring 3D cinemagraphy into the sight of a broader community and motivate further research.  

\noindent\textbf{Limitations and future work.}
Our method may not work well when the depth prediction module estimates erroneous geometry from the input image, e.g., thin structures. In addition, inappropriate motion fields will sometimes lead to undesirable results, e.g., some regions are mistakenly identified as frozen. As we take the first step towards 3D cinemagraphy, in this paper, we focus on handling common moving elements, i.e., fluids. In other words, our method may not apply to more complex motions, e.g., cyclic motion. We leave this for our future work. 

\noindent\textbf{Acknowledgements.}
This study is supported under the RIE2020 Industry Alignment Fund – Industry Collaboration Projects (IAF-ICP) Funding Initiative, as well as cash and in-kind contribution from the industry partner(s). This work is also supported by Adobe Gift and the Ministry of Education, Singapore, under its Academic Research Fund Tier 2 (MOE-T2EP20220-0007) and Tier 1 (RG14/22).

{\small
\bibliographystyle{ieee_fullname}
\bibliography{main}
}

\end{document}